\documentclass[runningheads]{llncs}

\usepackage[T1]{fontenc}
\def\doi#1{\href{https://doi.org/\detokenize{#1}}{\url{https://doi.org/\detokenize{#1}}}}
\usepackage{graphicx}
\usepackage{multirow}
\usepackage{booktabs}
\usepackage{makecell}
\usepackage{amssymb, amsmath, bm}
\usepackage{listings}
\begin{document}
\title{A Faster, Lighter and Stronger Deep Learning-Based Approach for Place Recognition}
\titlerunning{A Faster, Lighter and Stronger Approach for Place Recognition}
\author{Rui Huang \and
Ze Huang \and
Songzhi Su}
% %
\authorrunning{R. Huang et al.}
\institute{School of Information, Xiamen University, Xiamen, 361005, Fujian, China
\email{donmuv@163.com}}
\maketitle              % typeset the header of the contribution
\begin{abstract}
Visual Place Recognition is an essential component of systems for camera localization and loop closure detection, and it has attracted widespread interest in multiple domains such as computer vision, robotics and AR/VR. In this work, we propose a faster, lighter and stronger approach that can generate models with fewer parameters and can spend less time in the inference stage. 
We designed RepVGG-lite as the backbone network in our architecture, it is more discriminative than other general networks in the Place Recognition task. RepVGG-lite has more speed advantages while achieving higher performance. We extract only one scale patch-level descriptors from global descriptors in the feature extraction stage. Then we design a trainable feature matcher to exploit both spatial relationships of the features and their visual appearance, which is based on the attention mechanism.
Comprehensive experiments on challenging benchmark datasets demonstrate the proposed method outperforming recent other state-of-the-art learned approaches, and achieving even higher inference speed. 
Our system has 14 times less params than Patch-NetVLAD, 6.8 times lower theoretical FLOPs, and run faster 21 and 33 times in feature extraction and feature matching.
Moreover, the performance of our approach is 0.5\% better than Patch-NetVLAD in Recall@1. We used subsets of Mapillary Street Level Sequences dataset to conduct experiments for all other challenging conditions.

\keywords{Visual Place Recognition\and Computer Vision for Automation \and Deep Learning \and Augmented Reality \and Virtual Reality.}
\end{abstract}
\section{Introduction}
\label{intro}
Visual place recognition aims to help a robot or a vision-based navigation system determine whether it is at a previously visited location. In a visual Simultaneous Localization and Mapping (SLAM) system\cite{durrant2006simultaneous}, place recognition, which is also referred to as loop closure detection (LCD), is a key component. It can not only reduce the localization error induced by visual odometry (VO), but also avoid building an ambiguous map of the unknown environment\cite{newman2005slam}. 
VPR technology can build a map with a more realistic visual effect, assisting in enhancing the visual effect in the field of AR/VR.

Visual place recognition is the task of recognizing the place depicted in an image (or a sequence of images). This task is commonly referred to as an image retrieval problem~\cite{revaud2019learning}. All datasets have a series of images (database) that need to be retrieved. Each image in the database is tagged with a location identifier, such as a GPS coordinate. The place recognition system searches the database for images that are related to a new picture that needs to be localized (query). If related images are identified, the tagged locations of those images are utilized to infer the query's location.

Recently, deep learning has become one of the most attractive research topics and has achieved great success in many areas, including computer vision (CV), robotics and AR/VR. And there are two common ways to represent the query and database images: using global descriptors which describe the whole image~\cite{arandjelovic2016netvlad}~\cite{warburg2020mapillary}, or using local descriptors that describe areas of interest~\cite{dusmanu2019d2}~\cite{detone2018superpoint}.

\begin{figure}[t]
\centerline{\includegraphics[width=0.7\textwidth]{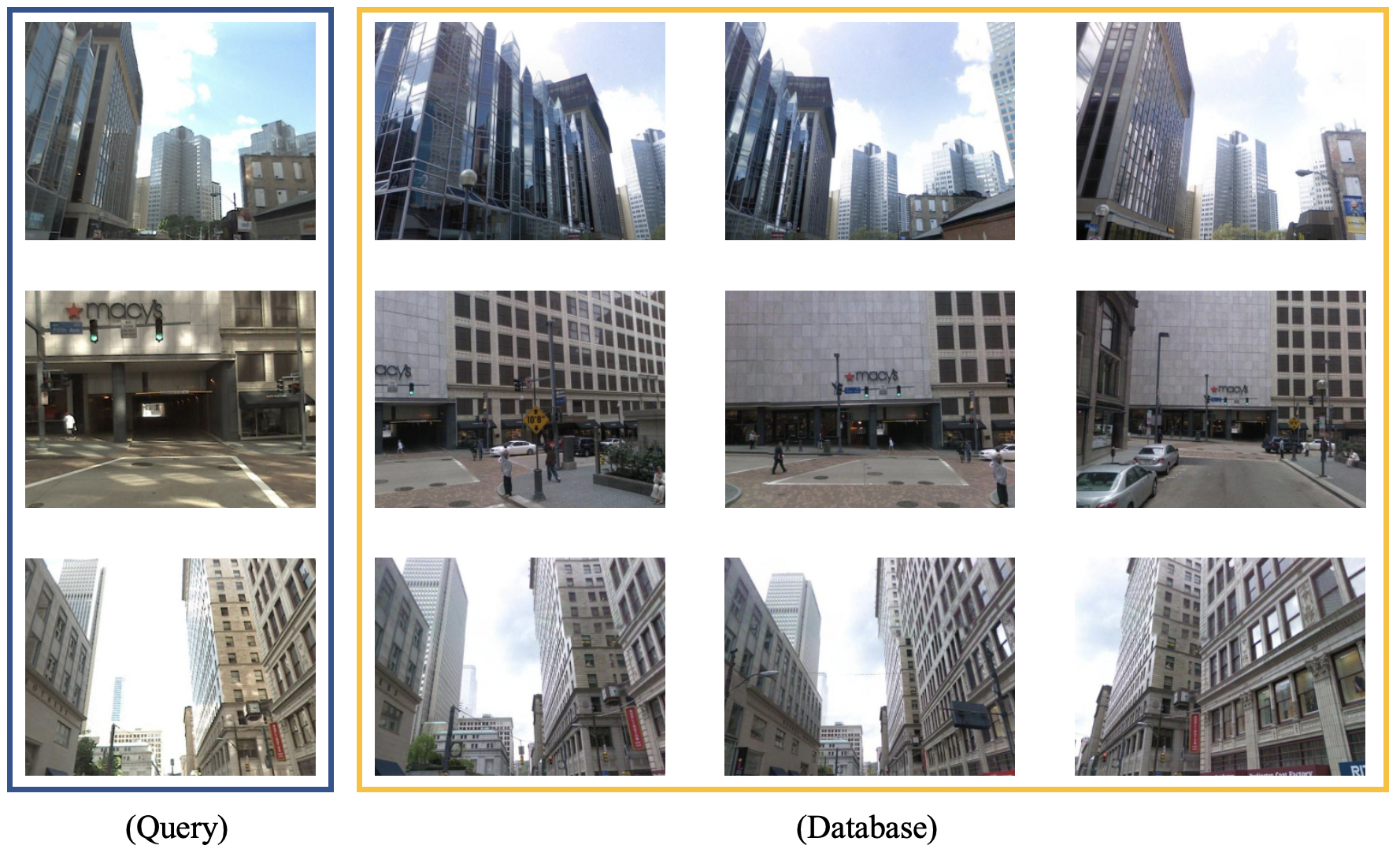}}
\caption{Examples of three queries corresponding to database in the Pittsburgh 30k dataset. In the three examples, the left column is query image, and the three columns on the right are database images. There are a large number of pictures that are taken from multiple perspectives in the database, a complex network may consume a lot of unnecessary time in retrieval work, may also result in incorrect matches.}
\label{fig:1}       % Give a unique label
\end{figure}

The novel Patch-NetVLAD~\cite{hausler2021patch} system proposed locally-global descriptors, these descriptors are extracted for densely-sampled local patches within the feature space using a VPR-optimized aggregation technique NetVLAD~\cite{arandjelovic2016netvlad}. And they presented a multi-scale fusion technique that generates and combines these hybrid descriptors of different sizes to achieve improved performance. But Patch-NetVLAD system produce a large model with more parameters, and it spend a lot of time in all stages. These problems will adversely affect the visual place recognition task, including detection efficiency and ease of deployment.

Transformer~\cite{vaswani2017attention} solely utilizes the attention mechanism to achieve superior quality in machine translation tasks, which dispense recurrence and convolutions entirely. Inspired by the success of the Transformer, we use self-attention to maximize both the spatial associations of the features and their visual appearance, and additionally introduces the cross-attention, which is symmetric. We derive patch features from the a global feature, NetVLAD, as the input of the attention layer.

In order to solve these problems, our contributions are summarized as follows.

First, we established a faster, lighter and stronger place recognition system in two stages. In the first stage, we design RepVGG-lite network to retain more local information. RepVGG-lite is a VGG-style architecture which outperforms many complicated models, it not only has stronger performance, but also has fewer parameters. Our method can have more speed advantages than Patch-NetVLAD while achieving higher performance. RepVGG-lite is a more discriminative network structure in the VPR task than other general networks.

Additionally, in the datasets related to the visual place recognition task, such as Pittsburgh 30k~\cite{torii2013visual}, Tokyo 24/7~\cite{torii201524}, Mapillary Street Level Sequences (MSLS)~\cite{warburg2020mapillary}, etc., their databases have many redundant images, which were taken in the same place at different angles or under different conditions (see Fig.~\ref{fig:1}, there are a lot of pictures in the database that are pretty similar). Datasets in cities are generally relatively static, with low diversity and little change. So, a complex network may consume a lot of unnecessary time in retrieval work, our system is not only simple but also robust and powerful. 

Second, to increase the speed of inference and strengthen the association between local information, we design a trainable attention-based feature matcher in the second stage, which consist of an attention module and an optimal transport layer. Then our system applied the assignment matrix generated by the optimal transport layer to re-order the images ranking list, resulting in the final image retrievals.

Last, extensive experiments demonstrate the effectiveness of our approach, including in the subtasks (e.g., day-night, Summer-Winter) of MSLS appearance changes dataset, and in other challenging benchmark datasets. Notably, the performance on Pittsburgh reaches a recall of 91.15\% at top 1 retrieval.

\begin{figure*}[t]
\includegraphics[width=1\textwidth]{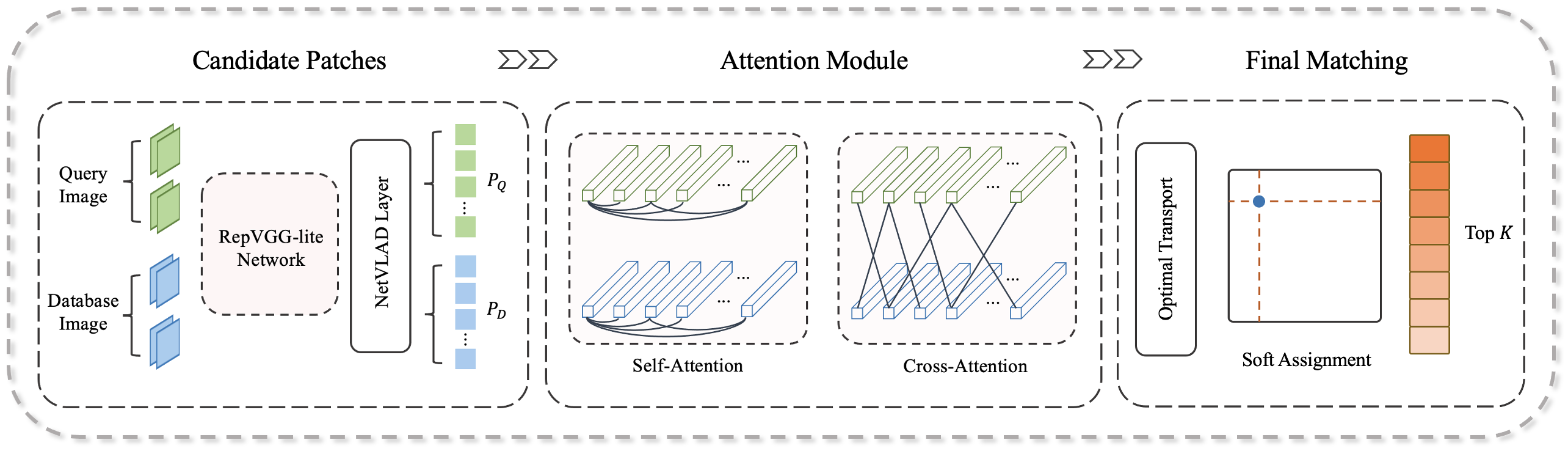}
\caption{\textbf{An overview of our proposed method.} We use RepVGG-lite as the backbone network in the first stage. We take a candidate patches list of most likely reference matches to a query image, ranked from NetVLAD global retrieval. In the second stage, we introduce self-attention and cross-attention to enhance the feature learning intra-image and inter-image. Then we use the optimal transport layer for computing confidence matrix so that get the final image matching at top $K$ retrieval.}
\label{fig2}
\end{figure*}

\section{The proposed algorithm}
Our system uses a more lightweight network RepVGG-lite as the backbone network in the first stage. And we use feature-map aggregation methods to extract descriptors for local patches within the whole feature map. We extract only one scale suitable patch-level descriptors. We use the attention module to compensate for the lack of local information caused by a single-scale patch in the second stage, then we solve the patches matching problem through an optimal transport layer. An overview of the complete pipeline can be found in Fig.~\ref{fig2}.

\subsection{RepVGG-lite Network}
NetVLAD~\cite{arandjelovic2016netvlad} network architecture leverages the Vector-of-Locally-Aggregated-Descriptors (VLAD) to build a condition and viewpoint invariant embedding of an image by aggregating the intermediate feature maps. Given $N$ $D$-dimensional local image descriptors $\{X_i\}_{i=1}^N$ as input, and $K$ cluster centers weighted by soft-assignment, the output VLAD image representation $V$ is $D\times K$-dimensional. The $(j,k)$ element of $V$ is computed as follows:
\begin{eqnarray}
V(j,k) = \sum_{i=1}^N a_k(X_i)(x_i(j)-c_k(j))
\end{eqnarray}
where $x_i(j)$ and $c_k(j)$ is the $j$-th dimension of the $i$-th descriptor and $k$-th cluster center, respectively, and $a_k(X_i)$ is 1 if cluster $c_k$ is the closest cluster to descriptor $X_i$ and 0 otherwise.

NetVLAD and some other systems employ VGG~\cite{simonyan2014very} as the backbone network, which result in large model size and a lot of time spent in inference-time, which has a negative impact in practical applications. However, recognition tasks must be completed in a timely manner on a computationally constrained platform in many real-world applications such as robots, self-driving cars, and augmented reality. So we need a lighter and more robust network. RepVGG~\cite{ding2021repvgg} has a RepVGG-A model with fewer convolutional layers and a RepVGG-B model with more convolutional layers, which have higher speed and higher performance respectively. We choose the smallest model RepVGG-A0 with 1280 output channels to adapt to the VPR task.

Furthermore, we cut out the last stage of RepVGG-A0, called RepVGG-lite, which is our system's backbone network with only 192 output channels. Specifically, the 1 × 1 branches of training-time RepVGG can be removed by structural re-parameterization, then the body of inference-time RepVGG only involves one single type of operation: 3 × 3 conv followed by ReLU, which makes RepVGG faster in inference-time. But in order to leave more local information and make the network more discriminative in the VPR task, we only prune the last layer without re-parameterization, which has little effect on our model. Table.~\ref{tab1} shows the specification of RepVGG-lite including the depth and width.

\renewcommand\arraystretch{1.3}
\begin{table}
\caption{Network Specification of RepVGG-lite}
\centering
\begin{tabular}{c|c|c|c} 
\hline
Stage  & Outputize & Layers & Channels of layer  \\
\hline
1  & 320$\times$240   & 1  & 48 \\
2  & 160$\times$120   &  2  &  48    \\
3  & 80$\times$60   & 4 & 96 \\
4  & 40$\times$30   & 14 & 192 \\
\hline
\end{tabular}
\label{tab1}
\end{table}

\subsection{Candidate Patches Extraction}
Inspired by the Patch-NetVLAD~\cite{hausler2021patch}, we also extract descriptors for intensively sampled sub-regions (in the form of patches) within the whole feature map. Our system extract a set of $d_x\times d_y$ patches $\{P_i,x_i,y_i\}$ with stride $s_p$ from the feature map, where the total number of patches is given by
\begin{equation}
n_p = \lfloor\frac{H-d_y}{s_p}+1\rfloor * \lfloor\frac{W-d_x}{s_p}+1\rfloor, d_y, d_x\le H, W
\end{equation}
The set of patch features and the coordinate of the patch's center inside the feature map are denoted by $P_i$ and $x_i,y_i$, respectively. 

Notably, in order to improve the speed of feature extraction, our system extract suitable patch-level descriptors at only one scale, which can also avoids over-retrieval to improve retrieval performance since the reference images in datasets contain a large number of redundant images.

Patch-NetVLAD compute the spatial matching score used for image retrieval, which employs a RANSAC-based scoring method. Finally Patch-NetVLAD applied these match scores to re-order the initial candidate list, resulting in the final image retrievals. However, using this RANSAC-based spatial matching score for image retrieval consumes a lot of time, on average 14557 ms per query image. So we introduced the attention mechanism improve retrieval efficiency and enhance the relationships of the features.

\begin{figure}[t]
\centerline{\includegraphics[width=0.8\textwidth]{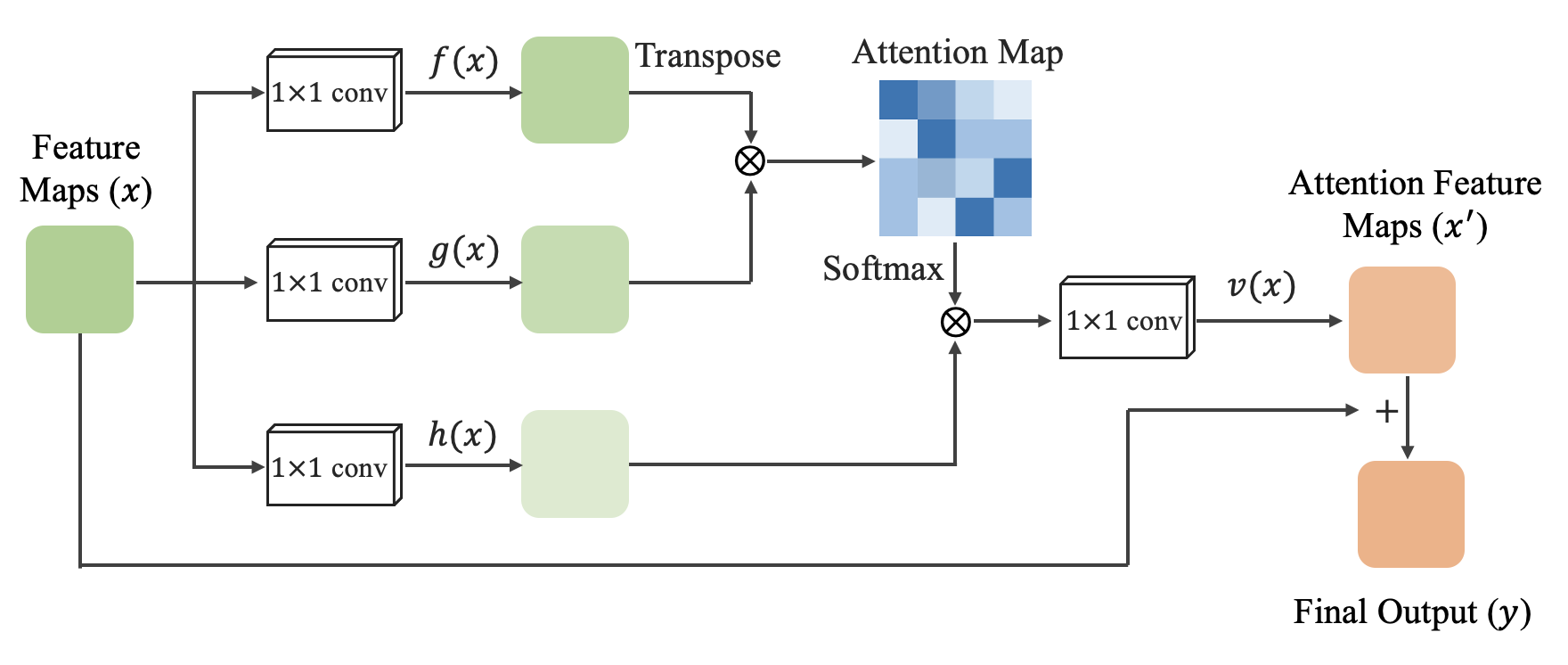}}
\caption{Illustration of the attention module, which can encode meaningful spatial relationships between local descriptors. The $\otimes$ denotes matrix multiplication, and the $+$ is matrix addition.}
\label{fig:3}       % Give a unique label
\end{figure}

\subsection{Attention Module}
Inspired by the success of ~\cite{vaswani2017attention}, we introduce an attention module to further process patches descriptors in the second stage, including a self-attention layer and a cross-attention layer. The self-attention module computes response at a position as a weighted sum of the features at all positions, where the weights (or attention vectors) calculated at a low computational cost. Fig.\ref{fig:3} presents its architecture. Given candidate patches descriptors $\bm{x}\in \mathbb{R}^{C\times N}$, where $C$ is the number of channels and $N$ is the number of patches descriptors from the previous network. We first transform $\bm{x}$ into two new feature spaces $\bm{f}$, $\bm{g}$ respectively to calculate the attention, where $\bm{f(x)=W_fx}$, $\bm{(x)=W_gx}$. Then the attention map $\bm{\rho}$ is calculated, defined as follows:
\begin{equation}
\rho_{j,i} = \frac{exp(f(x_i)^T\cdot g(x_j))}{\sum^N_{i,j=1} exp(f(x_i)^T\cdot g(x_j))}
\end{equation}
where $\rho_{j,i}$ indicates that the $i^{th}$ features impacts on $j^{th}$ features, with the shape of $N\times N$. The final output enhances feature integration by combining spatial geometric relationships and global contextual information. 

On the other hand, the cross-attention is symmetric to self-attention, which is critical for image matching. The cross-attention enables cross-image communication and is inspired by the way humans look back-and-forth when matching images.

\subsection{Final Matching}
The second major module of second stage (see Fig.\ref{fig2}) is the optimal matching layer~\cite{sarlin2020superglue}, which produces a soft assignment matrix. Each possible correspondence have a confidence value. The assignment $Z\in [0,1]^{M\times N}$ can be generated by calculating a score matrix $C\in \mathbb{R}^{M\times N}$ for all possible matches. Specifically, the score matrix $C$ represents the similarity of each pair of matching descriptors:
\begin{equation}
C_{i,j}=<y_i^Q, y_j^D>, \quad \forall (i,j)\in Q\times D
\end{equation}
where $<\cdot,\cdot>$ denotes the inner product, and we assume that the query image $Q$ and the database $D$ image have $M$ and $N$ patch descriptors, indexd by $Q:=\{1,...,M\}$ and $D:=\{1,...,N\}$, respectively. The matching descriptors are not normalized like learnt visual descriptors, and their value might change to reflect the prediction confidence during training. 

The optimal transport~\cite{peyre2019computational} can solve the above optimization problem. Its entropy-regularized formulation can yield the desired soft assignment, and then be solved efficiently on GPU using the Sinkhorn algorithm. Finally, our system applied the assignment matrix to re-order the initial candidate images ranking list, resulting in the final image retrievals.

\subsection{Loss Funtion}
Both the first and second stage models of our designed system are trainable. In the first stage, our model computes the loss using triplet ranking loss, which is consistent with NetVLAD~\cite{arandjelovic2016netvlad}. And in the second stage our system use the negative log-likelihood loss function to calculate the loss. Given the ground truth matches $G=\{(i,j)\}\subset Q\times D$ estimated from GPS/UTM coordinate of datasets, then we minimize the negative log-likelihood of the assignment $Z^{'}$:
\begin{equation}
Loss=-\sum\limits_{(i,j)\in G}log Z^{'}_{i,j}
\end{equation}

\section{Experimental Results}
Our system is trained separately on the Pittsburgh 30k~\cite{torii2013visual} dataset, Tokyo24/7 dataset~\cite{torii201524} and Mapillary Street Level Sequences dataset~\cite{warburg2020mapillary}. The scale of candidate patch in our system is square patch sizes 2. All datasets are evaluated using the Recall@$K$ metric, VPR techniques commonly form a part of localization pipelines for generating location priors. For a given localization radius, Recall@$K$ is defined as the ratio of correctly retrieved queries within the top $K$ predictions to the total number of queries.

\subsection{Datasets details}
To evaluate our approach, we used various outdoor key benchmark datasets: Pittsburgh~\cite{torii2013visual}, Tokyo24/7~\cite{torii201524} and Mapillary Streets~\cite{warburg2020mapillary}. Mapillary Street Level Sequences (MSLS) is a recently released dataset for place recognition. It consists of image sequences from a diverse set of cities (30 major cities across the globe), captured under a variety of appearance conditions. In our experiments, we used Melbourne for training and Amman, Trondheim, Goa and London for testing. Pittsburgh 30k contains 10k database images downloaded from Google Street View, which contains both appearance and viewpoint change. Tokyo24/7 dataset contains 76k database images and 315 query images taken using mobile phone cameras, we use the model trained on Pittsburgh to test on the Tokyo24/7 dataset.

\renewcommand\arraystretch{1.3}
\begin{table}[t]
\caption{Performance comparisons on Pittsburgh and Tokyo24/7 datasets. The R@K mean the Recall@K. The $\dagger$ denote $D_{PCA}=4096$, the $\ddagger$ denote $D_{PCA}=512$}
\centering
% \resizebox{\linewidth}{!}{
\begin{tabular}{c|c c c|c c c}
\multirow{2}{*}{Method} & \multicolumn{3}{c|}{Pittsburgh (Val.set)} & \multicolumn{3}{c}{Tokyo24/7} \\
\cline{2-7}
& R@1 & R@5 & R@10 & R@1 & R@5 & R@10 \\
\hline
\hline
AP-GEM~\cite{rublee2011orb}  & 78.51 & 92.92 & 95.76 & 40.33 & 56.12 & 66.34 \\
DenseVLAD~\cite{torii201524}  & 81.35 & 91.86 & 94.51 & 60.12 & 67.59 & 73.37 \\
NetVLAD~\cite{arandjelovic2016netvlad}  & 87.92 & 95.84 & 97.41 & 64.88 & 78.42 & 81.65 \\
SuperGlue~\cite{sarlin2020superglue}  & 90.59 & 97.01 & 98.23 & \textbf{88.24} & \textbf{90.21} & 90.28  \\
Patch-NetVLAD~\cite{hausler2021patch}$\dagger$  & 90.65 & 96.48 & 97.66 & 86.07 & 88.63 & 90.55 \\
Patch-NetVLAD~\cite{hausler2021patch}$\ddagger$  & 90.28 & 95.93 & 97.17 & 85.65 & 88.25 & 90.11 \\
\hline
\hline
\textbf{Ours}  & \textbf{91.15} & \textbf{97.38} & \textbf{98.52} & 86.93 & 89.07 & \textbf{90.61} \\
\end{tabular}
\label{tab2}
\end{table}

\begin{figure}[t]
\includegraphics[width=1.0\textwidth]{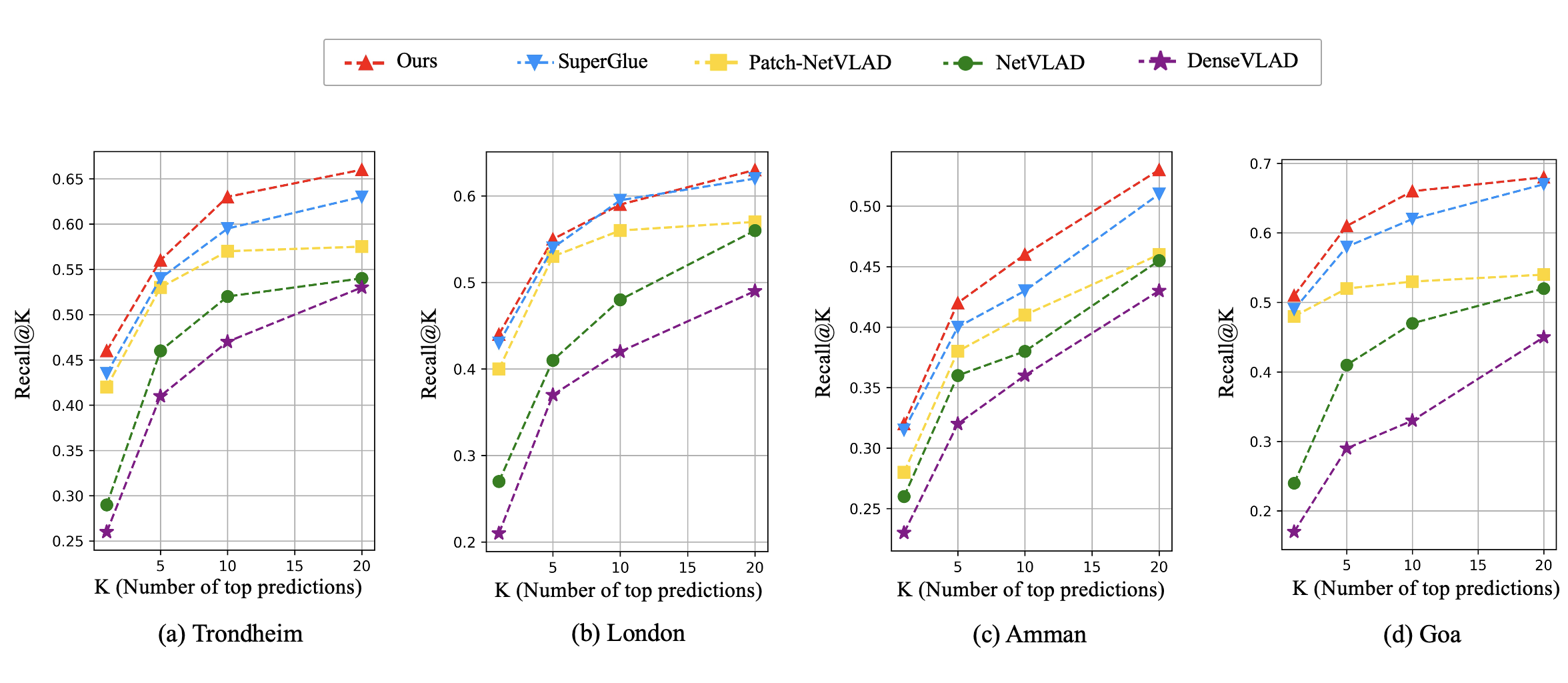}
\caption{Recall performance on MSLS dataset: (a) Trondheim, (b) London, (c) Amman and (d) Goa using a model trained on MSLS Melbourne.}
\label{fig:6}       % Give a unique label
\end{figure}

\subsection{Comparison to State-of-the-art Methods}
Table.~\ref{tab2} shows performance comparisons on Pittsburgh 30k dataset and Tokyo24/7 dataset for different method types including AP-GEM~\cite{rublee2011orb}, DenseVLAD~\cite{torii201524}, NetVLAD, Patch-NetVLAD methods and our system. Patch-NetVLAD~\cite{hausler2021patch} extract multi-scale patch-level descriptors from global features, which set square patch sizes 2, 5 and 8 with associated weights $w_i=0.45,0.15,0.4$ for the multi-scale fusion.

Then Patch-NetVLAD used the RANSAC scoring approach to compute a score matrix to re-order the initial candidate list. The $\dagger$ indicates the Patch-NetVLAD system with $D_{PCA}=4096$, which is the baseline of the Patch-NetVLAD method. Notably, since our model is lightweight and the output dimension is low, the $PCA$ dimensions of our system is set to 512. Nevertheless, the performance of our method still exceeds the Patch-NetVLAD system with $D_{PCA}=4096$. For fairness, we also show the experimental results of Patch-NetVLAD method with $D_{PCA}=512$, denoted as $\ddagger$, which has poor performance. 

Fig.~\ref{fig:6} shows recall performance on four cities of the MSLS dataset using a model trained on one city (Melbourne). MSLS offers much more variability in terms of weather, structure, viewpoint and domain changes~\cite{warburg2020mapillary}. Amman mainly has day-night variations, London and Trondheim mainly have old-new and summer-winter variations.

\begin{table}[t]
\caption{Comparative experiment with Patch-NetVLAD method in speed, params, model size and theoretical FLOPs. }
\centering
% \resizebox{\linewidth}{!}{
\begin{tabular}{c c c c c c}
\toprule
\multirow{2}{*}{Method} & Speed1 & Speed2 & Params & Theo & Meodel \\
& (ms) & (ms) & (M) & FLOPs (B) & Size (MB) \\
\hline
Patch-NetVLAD$\dagger$  & 625 & 14557 & 149.2 & 94.1 & 1100  \\
Patch-NetVLAD$\ddagger$  & 78 & 895 & 31.6  & 94.1  & 184  \\
\textbf{Ours}  & \textbf{31} & \textbf{438} & \textbf{5.3+5.1} & \textbf{8.5+5.3} & \textbf{32+10} \\
\bottomrule
\end{tabular}
\label{tab3}
\end{table}

Our method not only excels in retrieval accuracy, but also in inference speed. Table.~\ref{tab3} illustrates the advantages of our method over some other methods in terms of speed, model size, params and theoretical FLOPs. Where the Speed1 represents the time spent in extracting global features from images and generating candidate patch descriptors from the global feature map in the first stage, We calculate how many milliseconds it takes to process each image. The Speed2 indicates the time required for each query image to match the database images in the second stage, We calculate how many milliseconds it takes to process each query image to match. 

The experimental results show that our approach not only outperforms the best performing Patch-NetVLAD method, but also the speed of feature extraction and feature matching is very fast, about \textbf{21} times and \textbf{33} times that of Patch-NetVLAD system with $D_{PCA}=4096$, respectively. Although compared with the Patch-NetVLAD system with same $D_{PCA}=512$, our method is approximately 7.8 and 3.1 times faster in feature extraction and feature matching. Furthermore, we count the theoretical FLOPs, params and model size of Patch-NetVLAD and our methods. The calculation results show that the params of our system is about \textbf{14} times less than that of Patch-NetVLAD, theoretical FLOPs are about \textbf{6.8} times lower than it, and the model size is about \textbf{26} times smaller than it. 

Fig.~\ref{fig:7} compares partial query images retrieval results at top 1 of our method with the previous Patch-NetVLAD. In these examples, our proposed method successfully retrieves the matching reference image, while the novel Patch-NetVLAD produce incorrect place matches. 

\subsection{Comparison to Other Lightweight Models}
To highlight the performance and lightweight of our RepVGG-lite, we have conduct comparative experiments with other lightweight networks on Pittsburgh 30k dataset. These experiments only replace the backbone network in the first stage, and the rest of the settings are consistent with our baseline system. 

% % \renewcommand\arraystretch{1.3}
% \begin{table}[t]
% \caption{Our RepVGG-lite compare with other lightweight models on Pittsburgh dataset.}
% \centering
% % \resizebox{\linewidth}{!}{
% \begin{tabular}{c c c c c c}
% \toprule
% \multirow{2}{*}{Method} & \multirow{2}{*}{R@1} & Speed1 & Params & Theo & Model\\
% & & (ms) & (M) & FLOPs (B) & Size (MB)\\
% \hline
% SuperPoint~\cite{detone2018superpoint}  & 82.43 & 307 & \textbf{1.3} & 26.1 & 69  \\
% MobileNetV2~\cite{sandler2018mobilenetv2}  & 83.39 & 54  & 2.2 & 1.9 & 329  \\
% ShuffleNetV2~\cite{ma2018shufflenet} & 82.51 & 80  & \textbf{1.3} & \textbf{0.9} & 133  \\
% RepVGG-A0~\cite{ding2021repvgg} & 86.86 & 285 & 18.3 & 23.7 & 110 \\
% \textbf{RepVGG-lite}  & \textbf{91.15} & \textbf{31} & 5.3 & 8.5 & \textbf{32}  \\
% \bottomrule
% \end{tabular}
% \label{tab4}
% \end{table}

\begin{figure}[t]
\centerline{\includegraphics[width=0.8\textwidth]{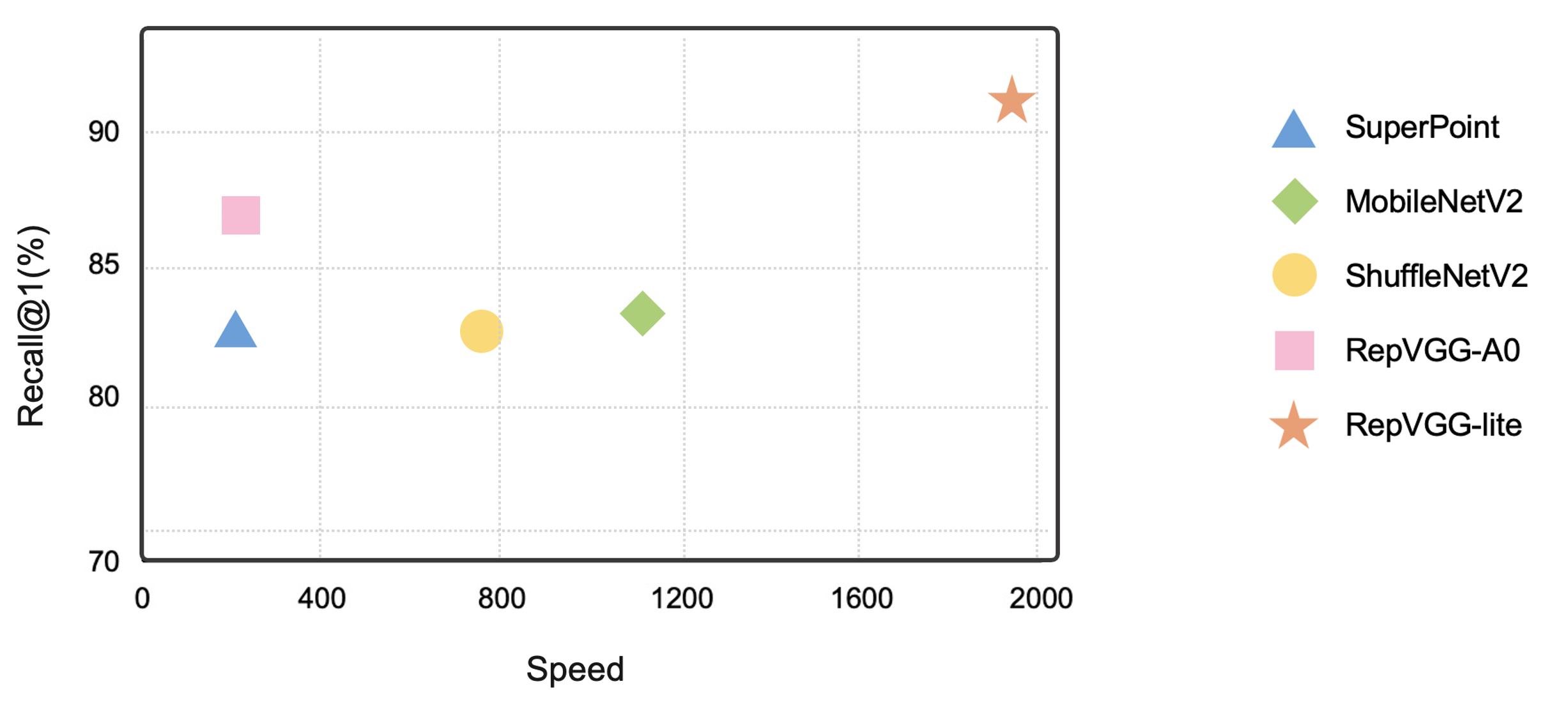}}
\caption{Comparison with other lightweight models. The Speed denote the number of images that can be processed per minute.}
\label{fig:4}       % Give a unique label
\end{figure}

As shown in Fig.~\ref{fig:4}, All models have added a PCA layer with 512 dimensions for fairness. The Speed in Fig.~\ref{fig:4} denote the number of images that can be processed per minute. In order to compare with SuperPoint~\cite{detone2018superpoint}, we modified it to extract global features in the first stage of our system. SuperPoint has a small model size and the least params, but its inference time is too long. Finally, we compared RepVGG-A0~\cite{ding2021repvgg} model, which is a little better than other lightweight networks in retrieval accuracy, but far from our RepVGG-lite model, and the RepVGG-A0 network has more params and theoretical FLOPs.

\section{Conclusion}
This paper demonstrates the power of our proposed RepVGG-lite and matcher in the first and second stages, respectively. The RepVGG-lite is a lighter and faster network, and it has significant performance gains, which is more discriminative than other general models in VPR task. The trainable matcher architecture uses two kinds of attention layer, self-attention and cross-attention. The matcher can enhances the receptive field of local descriptors, and supports cross-image communication. Comprehensive experiments demonstrate the proposed method achieves significant accuracy improvement over existing approaches, and achieving even higher inference speed, which is more realistic in the field of autonomous driving. 
We believe that the proposed method can provide considerable value to academia and industry in the robotics and AR/VR fields.

\begin{figure}[t]
\includegraphics[width=1\textwidth]{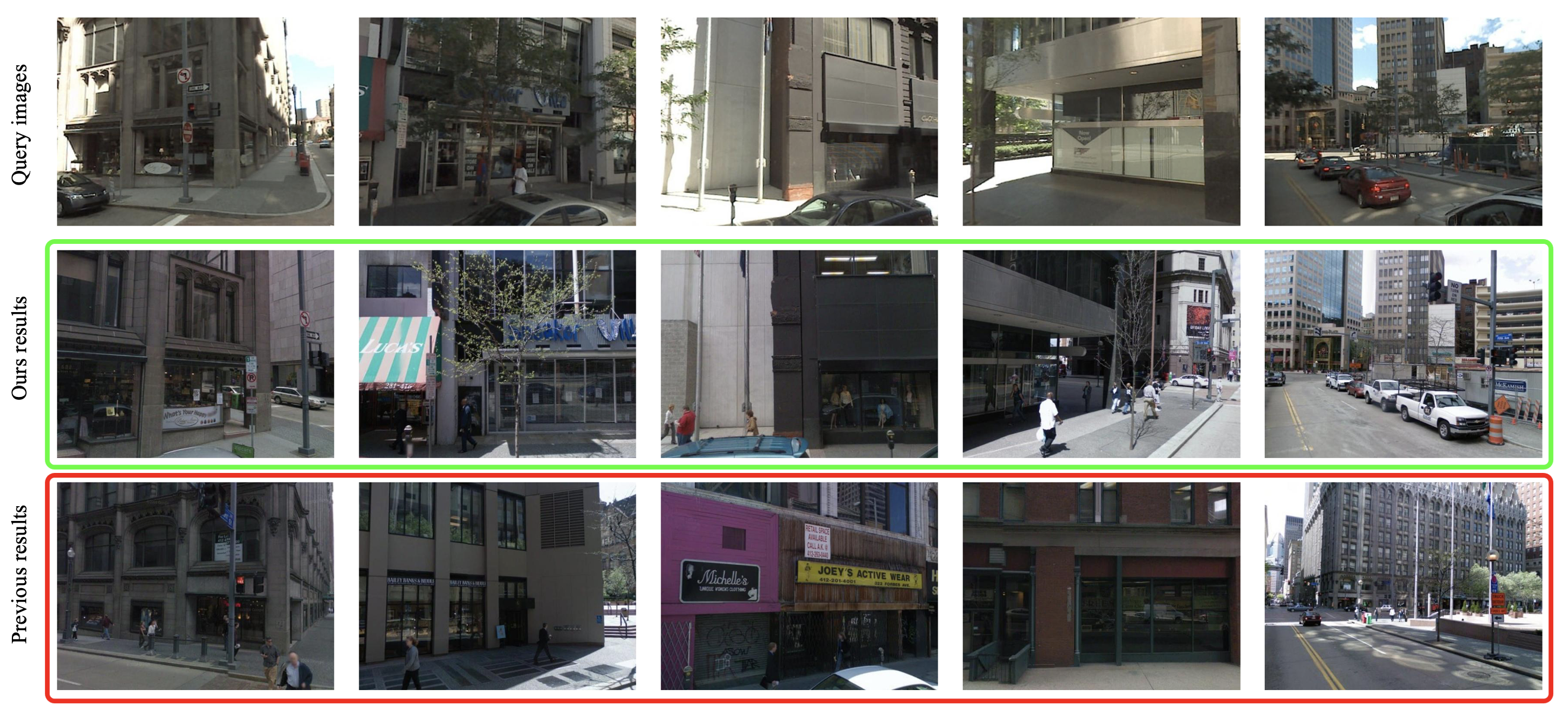}
\caption{Examples of retrieval results for challenging queries on Pittsburgh dataset. Each column corresponds to one test case: the query is shown in the first row, the top retrieved image using our proposed method in the second, and the top retrieved image using the previous Patch-NetVLAD in the last row. The green and red borders correspond to correct and incorrect retrievals, respectively.}
\label{fig:7}       % Give a unique label
\end{figure}



% \begin{table}
% % table caption is above the table
% \caption{Please write your table caption here}
% \label{tab:1}       % Give a unique label
% % For LaTeX tables use
% \begin{tabular}{lll}
% \hline\noalign{\smallskip}
% first & second & third  \\
% \noalign{\smallskip}\hline\noalign{\smallskip}
% number & number & number \\
% number & number & number \\
% \noalign{\smallskip}\hline
% \end{tabular}
% \end{table}

%\begin{acknowledgements}
%\end{acknowledgements}

% \bibliographystyle{spmpsci}      % mathematics and physical sciences
% \bibliographystyle{spphys}       % APS-like style for physics
\bibliographystyle{plain}
\bibliography{ref.bib}   % name your BibTeX data base

\end{document}